\documentclass{article}
\usepackage{ijcai25}

\usepackage{times}
\usepackage{soul}
\usepackage{url}
\usepackage[hidelinks]{hyperref}
\usepackage[utf8]{inputenc}
\usepackage[small]{caption}
\usepackage{graphicx}
\usepackage{amsmath}
\usepackage{amsthm}
\usepackage{booktabs}
\usepackage{algorithm}
\usepackage{algorithmic}
\usepackage[switch]{lineno}
\usepackage{amsmath,amssymb,amsfonts}
\usepackage{amsthm}
\usepackage{multirow}
\usepackage[hang, symbol, flushmargin]{footmisc}

\title{Morphology-optimized Multi-Scale Fusion: Combining Local Artifacts and Mesoscopic Semantics for Deepfake Detection and Localization
}

\author{
Chao Shuai$^{1,*}$\and
Gaojian Wang$^{1,*}$\and
Kun Pan$^{1,*}$\and
Tong Wu$^{1,*}$\and
Fanli Jin$^{1,*}$\and
Haohan Tan$^{1,*}$\and
Mengxiang Li$^{1,*}$\and
Zhenguang Liu$^{1,2}$\and
Feng Lin$^{1,2}$\And
Kui Ren$^{1,2}$\\
\affiliations
$^1$\small{The State Key Laboratory of Blockchain and Data Security, Zhejiang University}\\
$^2$\small{Hangzhou High-Tech Zone (Binjiang) Institute of Blockchain and Data Security}
\emails
\small{\{chaoshui, 12321142, pankun, cocotwu, 12421197, tanhh2023, limengxiang, flin, kuiren\}}@zju.edu.cn, liuzhenguang2008@gmail.com
}

\begin{document}

\maketitle

\footnotetext{* Equal contribution.}

\begin{abstract}
While the pursuit of higher accuracy in deepfake detection remains a central goal, there is an increasing demand for precise localization of manipulated regions. 
Despite the remarkable progress made in classification-based detection, accurately localizing forged areas remains a significant challenge.
A common strategy is to incorporate forged region annotations during model training alongside manipulated images. 
However, such approaches often neglect the complementary nature of local detail and global semantic context, resulting in suboptimal localization performance.
Moreover, an often-overlooked aspect is the fusion strategy between local and global predictions. 
Naively combining the outputs from both branches can amplify noise and errors, thereby undermining the effectiveness of the localization.

To address these issues, we propose a novel approach that independently predicts manipulated regions using both local and global perspectives. We employ morphological operations to fuse the outputs, effectively suppressing noise while enhancing spatial coherence. Extensive experiments reveal the effectiveness of each module in improving the accuracy and robustness of forgery localization.

\end{abstract}    
\section{Introduction}
\label{sec:intro}


The rapid advancement of deep generative models~\cite{goodfellow2014generative,ho2020denoising}, particularly deepfakes, has led to an unprecedented level of realism and sophistication in synthetic media. When such technologies are misused, the consequences are evident: the proliferation of misinformation, reputational damage, and the erosion of public trust. While existing deepfake detection models~\cite{hu2022finfer,shiohara2022detecting} have achieved notable progress in classification, their ability to pinpoint tampered regions remains limited. This deficiency hinders practical applications such as forensic analysis and content moderation, where identifying manipulated pixels is critical for accountability and trustworthiness.


Current deepfake localization methods exhibit three primary issues. First, \textit{local artifact detectors}, exemplified by CNN-based noise residual analysis~\cite{Hu2020SPAN} and edge inconsistency modeling~\cite{Yue2019ManTraNet}, suffer from semantic blindness, failing to detect contextually implausible manipulations. Second, \textit{global approaches} that leverage frequency spectrum analysis~\cite{Qian2020F3Net} or transformer-based contextual reasoning~\cite{Guillaro2023TruFor} lack the granularity to pinpoint fine-grained spatial anomalies. Third, hybrid frameworks~\cite{Wang2022ObjectFormer} that mechanically combine local/global features employ static fusion strategies, neglecting dynamic adaptation to manipulation characteristics across scales. Crucially, existing methods universally overlook \textit{mesoscopic artifacts}—manipulation traces manifesting at intermediate scales between pixel-level distortions and semantic contradictions~\cite{Zhu2025Mesoscopic}.

To tackle these challenges, we propose a novel morphology-optimized multi-scale fusion framework for deepfake detection and localization, which synergizes local forgery artifacts with mesoscopic semantic information. First, we present a two-stream Local Facial Forgery Detection and Location (\textbf{LFDL}) network that combines RGB and SRM (Steganalysis Rich Model) features through cross-modality consistency enhancement, effectively capturing fine-grained forensic traces like noise and edge artifacts. Second, we propose a Mesoscopic Image Tampering Localization (\textbf{MITL}) network that integrates frequency-enhanced representations with adaptive multi-scale weighting to encode both object-level and scene-consistency information. Third, we introduce a Morphology-Driven Mask Fusion (\textbf{MDMF}) strategy that intelligently merges LFDL and MITL localization results through differential dilation/erosion operations, generating reconciled global predictions.



Our key contributions are summarized as below:
\begin{itemize}
\item We propose a novel hybrid deepfake detection and localization framework that synergistically combines local and mesoscopic forgery cues.
\item We introduce a morphology-driven mask fusion strategy that adaptively refines localization masks by dilating local predictions and eroding mesoscopic predictions.
\item We evaluate the effectiveness and robustness of the proposed framework through extensive evaluation.
\end{itemize}  

\begin{table*}[h]
\caption{Supplementary Data Registration for Deepfake Detection Model Training: Model Types, Methods, and Forgery Details}
\centering
\begin{tabular}{ccccc}
\hline
Model Type                       & Method              & Forgery Types & Fake/Mask Image & Reference \\ \hline
\multirow{2}{*}{Image Edit}      & SBIs                & FaceSwap  &        18135    &       \cite{shiohara2022detecting}      \\
                                 & Random combination           & FaceSwap  &      17728            &     -        \\ \hline
\multirow{3}{*}{GAN}             & Simswap             & FaceSwap      & 14999             & \cite{chen2020simswap}       \\
                                 & MaskFaceGAN         & Face Attribute Editing              & 14999                 & \cite{pernuvs2023maskfacegan}              \\
                                 & Facedancer          & FaceSwap    & 20000             & \cite{rosberg2023facedancer}        \\ \hline
\multirow{2}{*}{Diffusion Model} & BELM                & Diffusion Inversion  & 14674        & \cite{wang2024belm}  \\
                                 & SD-inpanting        &  Inpanting             &  18347                 &  \cite{podell2023sdxl}           \\ \hline
\end{tabular}
\label{table:dataset}
\end{table*}

\section{Related Work}
\label{related_work}

\textbf{Deepfake Detection} Deepfake detection techniques have evolved significantly to counter the increasing sophistication of forgery paradigms. Current methods can be broadly categorized into four main approaches based on the cues they exploit: spatial domain \cite{liu2020global,nirkin2021deepfake,cao2022end,wang2023noise,tan2023learning}, temporal domain \cite{yang2019exposing,gu2022delving,yang2023masked,choi2024exploiting,peng2024deepfakes}, frequency domain \cite{qian2020thinking,li2021frequency,miao2022hierarchical,guo2023constructing,tan2024frequency}, and data-driven methods \cite{dang2020detection,zhao2021learning,hu2022finfer,huang2023implicit,guo2023hierarchical,zhang2024inclusion}. However, most existing methods focus solely on binary classification (real vs. fake), neglecting the localization of manipulated regions. This limitation not only restricts the practical utility of detection systems in forensic scenarios but also weakens model interpretability, as the lack of localization prevents both decision validation and forgery patterns analysis.




\noindent \textbf{Deepfake Location} Deepfake localization~\cite{miao2024mixture,miao2023multi,zhang2024mfms} aims to identify manipulated regions through pixel-level analysis. Existing methods for deepfake localization primarily focus on either local or global approaches to identifying manipulated regions. Local location methods, such as ManTra-Net \cite{Wu2019ManTraNet} and TruFor \cite{Guillaro2023TruFor}, detect anomalies within smaller regions of an image, focusing on fine-grained inconsistencies like noise and texture irregularities. Global location methods, including F3-Net \cite{qian2020thinking}, ObjectFormer \cite{Wang2022ObjectFormer}, and MVSS-Net \cite{chen2021MVSS}, consider the entire image or video frame, identifying larger-scale inconsistencies and manipulation artifacts that span broader areas. 

In contrast to existing methods, our approach integrates both local, mesoscopic, and global location information, applying morphological fusion techniques to enhance the robustness and precision of deepfake localization.
    
\section{Method}

\subsection{Overview}

\begin{figure*}[!t]
\centering
\centerline{\includegraphics[width=.9\textwidth]{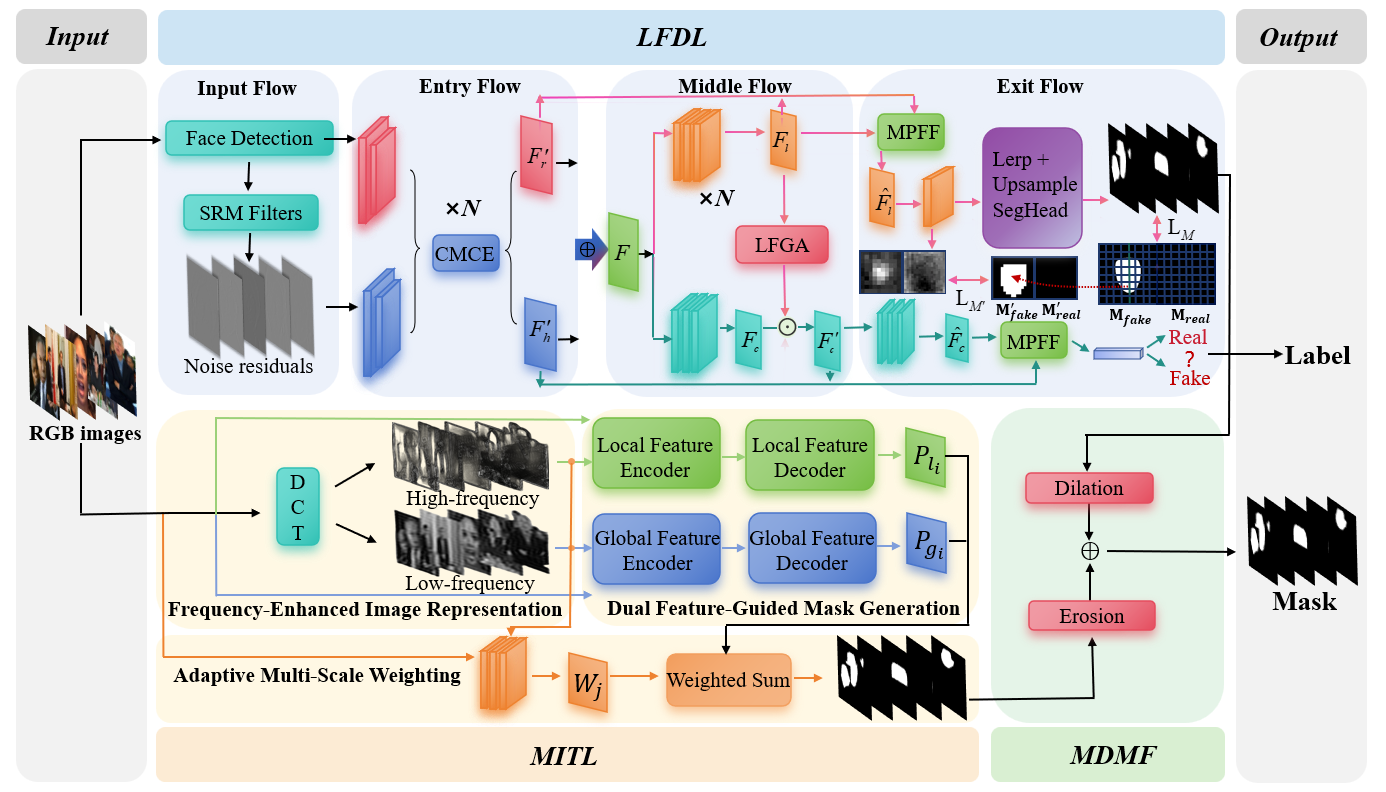}}
\caption{Overview of our proposed framework. (1) The \textbf{LFDL} module employs a two-stream architecture (RGB and SRM streams) and utilizes cross-modality consistency mechanisms for manipulation detection. (2) The \textbf{MITL} module processes frequency-enhanced features through dual encoders to achieve semantic-level tampering identification. (3) The \textbf{MDMF} module combines the dilated mask from \textbf{LFDL} (enhancing edge coherence) and the eroded mask from \textbf{MITL} (suppressing over-prediction) through a union operation to achieve comprehensive localization.}
\label{fig_over}
\end{figure*}

Our proposed framework is designed for robust deepfake detection and localization by synergizing local forgery artifacts with mesoscopic semantic information, to formulate a comprehensive and precise global prediction. The overall framework, as depicted in Figure~\ref{fig_over}, comprises three main components: the Local Facial Forgery Detection and Localization (LFDL) network, the Mesoscopic Image Tampering Localization (MITL) network, and the Morphology-Driven Mask Fusion (MDMF) strategy.

\subsection{Data Preparation}

\paragraph{1) Generating Diverse Forgery Images and Masks:}

To improve the model's robustness to diverse manipulation patterns, we construct a supplementary dataset by applying various forgery techniques to authentic images. Instead of relying on external datasets, we utilize the original real images to generate manipulated samples using representative methods spanning traditional image editing, GAN-based synthesis, and diffusion models. This process not only increases the diversity of forgery types but also provides paired masks for supervised learning. The details of the constructed dataset are shown in Table~\ref{table:dataset}. In total, 118,882 manipulated images were generated using various forgery techniques, each accompanied by a corresponding mask.

\paragraph{2) Data Preprocessing Strategy:}

To tackle the challenging task of facial forgery detection, we employ two distinct models, each is designed to analyze different forgery scales (e.g., facial attribute and background context), respectively. Correspondingly, two preprocessing pipelines are adopted to prepare the input for these models, ensuring alignment with their respective detection strategies. The LFDL module focuses on analyzing local facial regions, and its preprocessing pipeline emphasizes face localization and adaptive cropping to highlight detailed facial features. In contrast, the MITL module leverages global image context, and its preprocessing approach uses the entire original image and corresponding forgery mask without any facial cropping. By pairing these models with tailored preprocessing techniques, we simultaneously utilize localized and global cues, enhancing the robustness and comprehensiveness of the overall detection system.

In the LFDL module, face detection is applied to localize the face region in the image. If the detected face occupies a moderate portion of the entire image, the face region is cropped; otherwise, the original image is fed directly into the model. This adaptive cropping strategy highlights local facial details while preventing excessive cropping of small face regions, thereby preserving key information.

In contrast, the MITL module avoids face detection or cropping altogether. Instead, the entire original image and its corresponding forgery mask are directly input into the model. By preserving both global structure and background, this method leverages the overall context of the image in addition to facial cues. This preprocessing pipeline emphasizes global information, complementing the localized focus of LFDL, and together they enhance the completeness and robustness of facial forgery detection.

\subsection{LFDL: Local Facial Forgery Detection and Location with Two-Stream Architecture}

Deepfake manipulations often introduce subtle and spatially confined artifacts in key facial regions, such as the eyes, mouth, and contours. These fine-grained inconsistencies are difficult to capture using coarse global features alone. To address this, we employ a patch-based strategy wherein facial regions are first detected and cropped, then processed by a dedicated Two-Stream Network~\cite{shuai2023locate} designed to capture both RGB image and SRM noise residuals forgery cues at a local level.

The architecture comprises two parallel branches, classification and localization, each augmented by four key components that jointly exploit cross-modal consistency and patch-level context, as described below.

\paragraph{1) Cross-Modality Consistency Enhancement (CMCE):}

Let \(F_{\text{rgb}}^l \in \mathbb{R}^{C_l \times H_l \times W_l}\) and \(F_{\text{srm}}^l \in \mathbb{R}^{C_l \times H_l \times W_l}\) denote the features extracted from the \(l\)-th layer of the RGB and SRM streams, respectively. CMCE computes a cross-modal consistency map by measuring the cosine similarity at each spatial location:

\begin{equation}
\text{Corr}(f_i^r, f_i^h) = \frac{f_i^r \cdot f_i^h}{\|f_i^r\|_2 \|f_i^h\|_2}, \quad i \in \{1, \dots, H_l W_l\}
\end{equation}

where \(f_i^r, f_i^h \in \mathbb{R}^{C_l \times 1 \times 1}\) are the corresponding spatial vectors from the RGB and SRM streams. This correlation is used to refine both modalities:

\begin{equation}
F_r' = \text{ReLU}(F_r + \text{Corr} \odot F_h), \quad
F_h' = \text{ReLU}(F_h + \text{Corr} \odot F_r)
\end{equation}

Finally, the enhanced representation is obtained by summing the two refined features:

\begin{equation}
F_{\text{cmce}}^l = F_r' + F_h'
\end{equation}

This mechanism reinforces cross-modal interactions, allowing the network to exploit complementary forgery clues present in both spatial and frequency domains.

\paragraph{2) Local Forgery Guided Attention (LFGA):}

To mitigate the common failure case where networks overly rely on unmanipulated regions, LFGA explicitly guides attention toward potentially tampered areas. It first constructs a self-attention map from the localization branch’s feature \(F_l\) as:

\begin{equation}
\text{Att}_{ij} = \text{Softmax}(g(F_l)_i \cdot g(F_l)_j)
\end{equation}

where \(g(\cdot)\) is a learnable linear projection and \(i,j\) index spatial locations. This attention map highlights patch-wise similarity and forgery salience. The attention is then used to recalibrate the classification feature \(F_c\):

\begin{equation}
F_c^* = \text{ReLU}(\text{Reshape}(h(F_c) \otimes \text{Att}) + F_c)
\end{equation}

where \(h(\cdot)\) is a projection function and \(\otimes\) denotes matrix multiplication. LFGA is applied at multiple scales to progressively refine spatial focus and promote forgery-aware classification.

\paragraph{3) Multi-Scale Patch Feature Fusion (MPFF):}

Forgery traces often vary in spatial granularity. MPFF fuses features across multiple resolution levels to capture both coarse and fine-level inconsistencies. For the localization stream, high-resolution intermediate features \(F_{ml} \in \mathbb{R}^{c_2 \times h_2 \times w_2}\) and low-resolution features \(F_l \in \mathbb{R}^{c_1 \times h_1 \times w_1}\) are partitioned into corresponding non-overlapping patches. Within each patch \(P_k\), an intra-patch consistency is computed as:

\begin{equation}
f_{ml}^{(k,j)} = \text{Tanh}\left(\frac{\theta(p_k^j) \cdot \theta(f_l^k)}{c}\right)
\end{equation}

where \(\theta(\cdot)\) is a shared projection and \(c\) is a normalization constant. The same strategy is applied in the classification stream to generate \(f_{mc}^{(k,j)}\), allowing joint multi-scale fusion and consistency modeling. This patch-wise alignment helps bridge semantic gaps across scales while preserving spatial fidelity.

\paragraph{4) Semi-Supervised Patch Similarity Learning (SSPSL):}

Most public deepfake datasets lack pixel-level annotations of manipulated regions, which hinders effective supervision of the localization branch. To overcome this, SSPSL adopts a semi-supervised approach to generate pseudo ground-truth masks based on patch similarity analysis.

First, facial landmarks are used to detect the nose position, around which a rectangular area is heuristically designated as the manipulated region. Patches from this area are treated as pseudo-forged samples, while patches from authentic images serve as real references. 

Given the feature map \(F_f \in \mathbb{R}^{C \times H \times W}\), we compute cosine similarity between each patch feature \(f_{ij}^f\) and the average features of real (\(f_r\)) and forged (\(f_a\)) patches:

\begin{equation}
S_{ij}^{fr} = \frac{f_{ij}^f \cdot f_r}{\|f_{ij}^f\|_2 \|f_r\|_2}, \quad
S_{ij}^{ff} = \frac{f_{ij}^f \cdot f_a}{\|f_{ij}^f\|_2 \|f_a\|_2}
\end{equation}

Each patch is then assigned a binary label indicating whether it is more similar to real or forged examples:

\begin{equation}
M_{ij} = 
\begin{cases}
0, & S_{ij}^{fr} - S_{ij}^{ff} \geq 0 \\
1, & S_{ij}^{fr} - S_{ij}^{ff} < 0
\end{cases}
\end{equation}

This results in a pseudo mask \(M\) that approximates manipulated regions, which is further converted into patch-level labels. For a patch \(P_k\), its label \(M_k\) is defined by the average value of its corresponding pixels:

\begin{equation}
M_k = 
\begin{cases}
0, & \text{avg}(M_{P_k}) = 0 \\
1, & \text{avg}(M_{P_k}) > 0
\end{cases}
\end{equation}

The localization branch is supervised using a binary cross-entropy loss between predicted mask logits \(\hat{M}_k\) and pseudo labels \(M_k\):

\begin{equation}
\mathcal{L}_{\text{loc}} = -\frac{1}{h_1 w_1} \sum_{k=1}^{h_1 w_1} \left[ M_k \log \hat{M}_k + (1 - M_k) \log (1 - \hat{M}_k) \right]
\end{equation}

\noindent Meanwhile, the classification branch is optimized with a standard binary cross-entropy loss:

\begin{equation}
\mathcal{L}_{\text{cls}} = -\left[ y \log \hat{y} + (1 - y) \log (1 - \hat{y}) \right]
\end{equation}

\noindent \textbf{Overall Training Objective:} The total loss integrates both classification and localization terms for joint optimization:

\begin{equation}
\mathcal{L}_{\text{total}} = \mathcal{L}_{\text{cls}} + \mathcal{L}_{\text{loc}}
\end{equation}

This formulation allows the network to learn fine-grained forgery localization without explicit ground-truth annotations, enabling more scalable and flexible training under semi-supervised settings.

%

\subsection{MITL: Mesoscopic Image Tampering Localization with Mesorch}
Existing image tampering localization methods primarily rely on microscopic features, such as image RGB noise, edge signals, or high-frequency features, to detect tampering traces. However, these methods fail to effectively capture the macroscopic semantic information of tampering, leading to insufficient localization accuracy in complex scenes and difficulty in handling tampering related to semantics. To address this challenge, we apply the Mesorch~\cite{Zhu2025Mesoscopic} architecture to extract both microscopic details and macroscopic semantic information, and dynamically adjust the importance of different features through an adaptive weighting module. 

The framework mainly consists of three parts: Frequency-Enhanced Image Representation, Dual Feature-Guided Mask Generation, and Adaptive Multi-Scale Weighting, as described below.


\paragraph{1) Frequency-Enhanced Image Representation (FEIR):} 
The FEIR module initiates spectral decomposition using Discrete Cosine Transform (DCT) to amplify mesoscopic-level artifacts. Given an input RGB image \(x \in \mathbb{R}^{H \times W \times 3}\), we decompose it into high-frequency (\(x_h\)) and low-frequency (\(x_l\)) components through DCT filtering, retaining spatial dimensions \(H \times W \times 3\) for both components. These frequency-enhanced features are concatenated with the original image along the channel axis to form enhanced representations:

\begin{equation}
\begin{aligned}
I_h &= \text{Concat}(x, x_h) \in \mathbb{R}^{H \times W \times 6}, \\
I_l &= \text{Concat}(x, x_l) \in \mathbb{R}^{H \times W \times 6}
\end{aligned}
\end{equation}

\(I_h\) and \(I_l\) are enhanced representations used for further processing. This hybrid representation bridges pixel-level anomalies (microscopic) with semantic contradictions (macroscopic), enabling the detection of subtle manipulation traces that span hierarchical levels.

\paragraph{2) Dual Feature-Guided Mask Generation (DFGMG):}
This module is a key component of the whole architecture. It processes high-frequency and low-frequency enhanced images separately to capture fine-grained details and macroscopic semantics, respectively. The outputs from these encoders are then decoded to generate initial predictions, which will be combined in the next module to produce the final mask.

The high-frequency enhanced image \(I_h\) is processed by the Local Feature Encoder, a CNN-based architecture specifically engineered to capture fine-grained local details. These details play a critical role in the detection of microscopic artifacts within the image. Concurrently, the low-frequency enhanced image \(I_l\) is processed by the Global Feature Encoder, a Transformer-based framework designed to extract macroscopic semantics and global contextual information. Such information is essential for comprehending object-level manipulations in the image.

Each encoder outputs feature maps at four distinct scales:
\begin{equation}
\begin{aligned}
\{L_{s_1}, L_{s_2}, L_{s_3}, L_{s_4}\} &= \text{LocalFeatureEncoder}(I_h), \\
\{G_{s_1}, G_{s_2}, G_{s_3}, G_{s_4}\} &= \text{GlobalFeatureEncoder}(I_l) \\
\end{aligned}
\end{equation}

where \(C_{\text{local}}\) and \(C_{\text{global}}\) denote the total number of output channels at each scale \(i\) for the local and global encoders, respectively.
These feature maps are then processed by the corresponding decoders to generate initial predictions of manipulated regions:
\begin{equation}
\begin{aligned}
M_{l_i} &= \text{LocalFeatureDecoder}(L_{s_i}), \\
M_{g_i} &= \text{GlobalFeatureDecoder}(G_{s_i})
\end{aligned}
\end{equation}

where \(M_{l_i}\) and \(M_{g_i}\) are the local and global prediction masks, respectively, with a shape of \(H/4 \times W/4 \times 1\) for each scale \(i\).

\paragraph{3) Adaptive Multi-Scale Weighting (AMW):}
This Module dynamically optimizes the fusion of multi-scale representations by adjusting the importance of features across different scales. Unlike traditional methods that assume equal weighting, this module identifies and emphasizes critical scales while suppressing redundant or noisy information, thereby enhancing localization precision and robustness.  

The module takes three inputs: the original RGB image \( x \in \mathbb{R}^{H \times W \times 3} \), high-frequency components \( x_h \in \mathbb{R}^{H \times W \times 3} \), and low-frequency components \( x_l \in \mathbb{R}^{H \times W \times 3} \), both derived via DCT. These inputs are concatenated to form a composite representation \( I_{\text{concat}} = \{x, x_h, x_l\} \in \mathbb{R}^{H \times W \times 9} \).  

Then the weighting network will produce a tensor \( W \in \mathbb{R}^{\frac{H}{4} \times \frac{W}{4} \times 8} \), where each element reflects the relative importance of predictions from local and global features across four scales (totaling eight branches). Formally:  
\begin{equation}
W = \text{WeightingModule}(I_{\text{concat}})
\end{equation} 

The final prediction mask \( M \) is computed through pixel-wise weighted fusion of concatenated multi-scale predictions \( M_{\text{merge}} \in \mathbb{R}^{\frac{H}{4} \times \frac{W}{4} \times 8} \), followed by upsampling to the original resolution: 
\begin{equation}
M_{\text{merge}} = \text{Concat}(M_{l_1}, M_{g_1}, M_{l_2}, M_{g_2}, M_{l_3}, M_{g_3}, M_{l_4}, M_{g_4}) \\
\end{equation}

\begin{equation}
M = \text{Resize}\left( \sum_{j=1}^{8} W_j \odot M_{\text{merge}_j}, H, W \right)
\end{equation}

where \( \odot \) denotes element-wise multiplication. This adaptive fusion ensures spatial coherence while focusing on regions critical for mesoscopic artifact detection.




\begin{figure*}[!t]
\centering
\centerline{\includegraphics[width=0.7\textwidth]{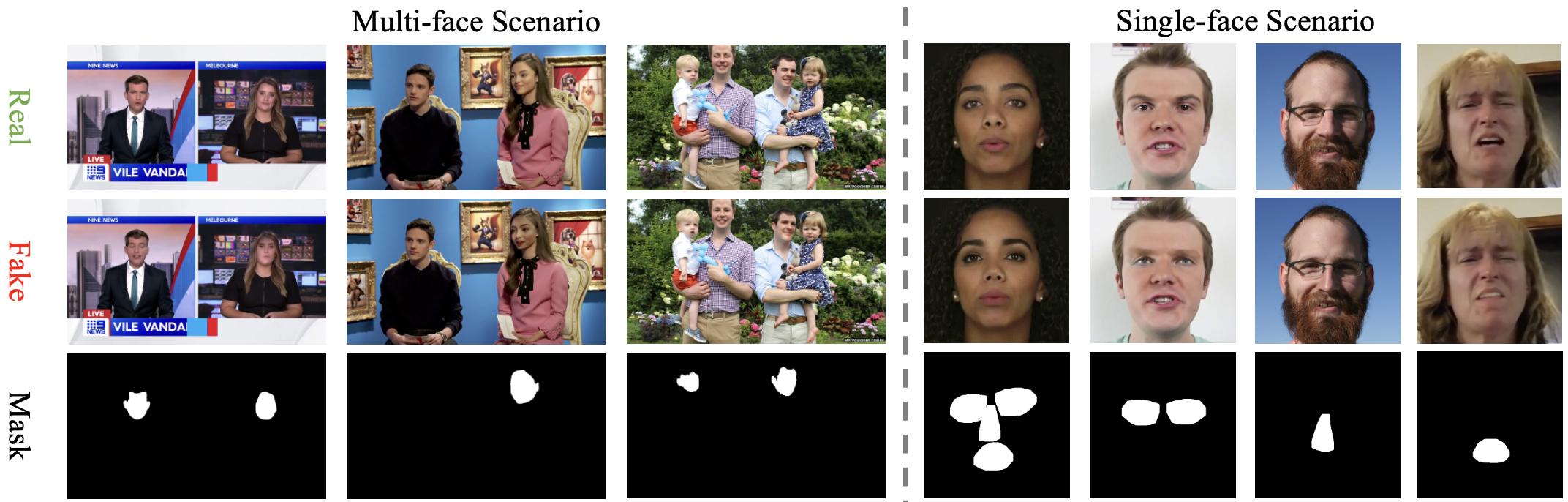}}
\caption{Examples of the DDL-I dataset.Multi-Face Scenario refers to an image with multiple faces where one or more faces have been altered. Single-Face Scenario involves altering a local region within an image that contains only one face.}
\label{fig_dataset}
\end{figure*}

\subsection{MDMF: Morphology-Driven Mask Fusion for Comprehensive Forgery Localization}

To leverage the complementary strengths of LFDL and MITL, we introduce a Morphology-Driven Mask Fusion strategy that combines their outputs for more accurate localization. 

The LFDL module first detects and crops the facial region, then performs fine-grained forgery localization. While effective in identifying facial manipulations, the cropping process may discard edge information, making the model sensitive to local artifacts. As a result, the output masks may have irregular boundaries or fragmented areas. Based on our observation that ground truth masks are generally smooth and coherent, we apply a dilation operation to the LFDL mask $M_{\text{LFDL}}$ to smooth edges and connect nearby manipulated regions:

\begin{equation}
M_{\text{LFDL}} \oplus B = \{ z \in \mathbb{Z}^2 \mid (B)_z \cap M_{\text{LFDL}} \neq \emptyset \}
\end{equation}

$B \subseteq \mathbb{Z}^2$ is a structuring element, here we set it as a $5 \times 5$ matrix of ones. $(B)_z = \{b + z \mid b \in B\}$ represents the translation of $B$ by displacement $z$. 

In contrast, the MITL module directly detects manipulated regions on the full image, effectively capturing non-facial forgeries such as hair alterations. However, the image resizing process during training may lead to loss of fine details, potentially resulting in ambiguous and over-extended predictions. To reduce this effect, we apply an erosion operation to the MITL mask $M_{\text{MITL}}$, which helps suppress over-prediction while retaining mesoscopic perception capabilities:

\begin{equation}
M_{\text{MITL}} \ominus B = \{ z \in \mathbb{Z}^2 \mid (B)_z \subseteq M_{\text{LFDL}} \}
\end{equation}

Finally, we take the union of the two processed masks to obtain the final localization result $M_{\text{final}}$:

\begin{equation}
M_{\text{final}} = (M_{\text{LFDL}} \oplus B) \cup (M_{\text{MITL}} \ominus B)
\end{equation} 

This morphology-based fusion compensates for the limitations of both modules: dilation recovers LFDL's lost edge coherence, while erosion suppresses MITL's over-prediction tendencies. Their union combines fine-grained localization with consistency analysis, yielding more accurate manipulation detection.

\section{Experiments}
\subsection{Experimental setup} 

\paragraph{Metrics.}
The performance of our model is evaluated using several key metrics. For detection, we report Area Under the ROC Curve (AUC). For spatial localization, we employ the F1 Score and Intersection over Union (IoU) to assess the accuracy of the localized tampered regions.

The F1 Score balances precision and recall, where precision measures the accuracy of predicted tampered pixels, and recall reflects the completeness of detection. It is computed as follows:

\begin{equation}
    F1\text{-}Score = \frac{2 \cdot \text{Precision} \cdot \text{Recall}}{\text{Precision} + \text{Recall}}
\end{equation}

Where:
\begin{equation}
    \text{Precision} = \frac{TP}{TP + FP}, \quad \text{Recall} = \frac{TP}{TP + FN}
\end{equation}

Here, \(TP\) denotes the correctly predicted tampered pixels, \(FP\) represents incorrectly predicted tampered pixels, and \(FN\) refers to missed tampered pixels.

IoU measures the overlap between the predicted and ground truth manipulation regions. It is calculated as:
\begin{equation}
    IoU = \frac{\text{Area of Intersection}}{\text{Area of Union}}
\end{equation}
or equivalently
\begin{equation}
    IoU = \frac{TP}{TP + FP + FN}
\end{equation}
Finally, the overall performance is summarized by a Final Score, which is the average of the three metrics:
\begin{equation}
    \text{Final score} = (\text{AUC} + \text{IoU} + \text{F1}) / 3
\end{equation}

Final score provides a comprehensive evaluation by combining detection and localization performance into a single metric.

\paragraph{Implementation Details.} 
Our approach consists of two primary models: LFDL and MITL, which are trained separately. The LFDL model utilizes an Xception backbone and is trained on 8 $\times$ A100 GPUs with a learning rate of $5\times10^{-4}$, a batch size of 76, and for 30 epochs. The MITL model, which combines a Segformer-B3 and ConvNeXt-Tiny hybrid backbone, is trained on 8 $\times$ RTX 4090 GPUs. It is trained with a learning rate of $10^{-4}$, a batch size of 10, and for 30 epochs. Following training, the models are fused during the inference stage to produce the final prediction. This modular training strategy allows each model to optimize performance by leveraging appropriate hardware for its own training, resulting in improved performance and efficiency.

\paragraph{Dataset.}
We evaluate our model on the Deepfake Detection and Localization Image (DDL-I) dataset~\cite{miao2025ddl}, which contains over 1.5 million samples with pixel-level annotations. DDL-I covers 61 latest deepfake methods across four forgery types: face swapping, face reenactment, full-face synthesis, and face editing. It encompasses both single-face and multi-face scenarios, providing a diverse range of forgery contexts. Additionally, pixel-level forgery region masks are provided, enabling precise localization of tampered areas.

\subsection{Experimental results} 

\begin{table*}[ht]
\centering

\caption{
Ablation Study. LFDL is a two-stream architecture focusing on local feature-based detection. MITL leverages the Mesorch backbone to extract both local artifacts and global semantic features. Mask Naive Fusion directly adds the output masks from LFDL and MITL, treating a pixel as fake if either branch flags it. MDMF further refines this fusion using morphological operations to enhance coherence and suppress noise. The final score is the average of AUC, F1-score, and IoU.
}

\begin{tabular}{lcccc}
\hline
Method                                              & Detection AUC & F1-score & IoU    & Final Score \\ \hline
LFDL                                          & 0.9790        & 0.6840   & 0.5981 & 0.7497      \\
MITL                                        & -             & -        & -      & 0.2349      \\
LFDL + MITL                           & -             & -        & -      & 0.3200      \\
LFDL + MITL + Mask Naive Fusion       & 0.9790        & 0.7598   & 0.6657 & 0.8015      \\
LFDL + MITL + MDMF & 0.9790        & 0.7759   & 0.6902 & 0.8150      \\ \hline
\end{tabular}
\label{tab:ablation} 
\end{table*}
Table~\ref{tab:ablation} presents the results of our ablation study, aimed at evaluating the contribution of different modules and fusion strategies in the proposed framework. We report Detection AUC, F1-score, IoU for localization accuracy, and a Final Score that aggregates the classification and localization quality.

LFDL is a patch-based forgery detection and localization module, which performs both binary classification and spatially localizes forgery artifacts. It focuses on fine-grained and local inconsistencies, especially in manipulated facial regions. Despite its localized input, it achieves strong performance with an AUC of 0.9790 and a respectable IoU of 0.5981, indicating its ability to precisely detect local manipulations.

MITL represents a standalone end-to-end deepfake detection model that takes the whole image as input and produces a full-resolution forgery mask. This model provides coarse but global localization information. When used alone, it yields a significantly lower Final Score of 0.2349, possibly due to lack of fine-grained localization capacity.

LFDL + MITL corresponds to a naive combination strategy where the detection confidence is derived from the LFDL module, while the global mask from the  MITL module is directly used for localization. This combination improves the Final Score to 0.3200, suggesting that LFDL's strong classification signal can benefit global prediction. However, without careful integration, the spatial coherence remains limited.

LFDL + MITL + Mask Naive Fusion applies a simple yet effective fusion strategy: the final localization mask is obtained by a pixel-wise logical OR operation between the LFDL module's patch-based local mask and the MITL module's global mask. This ensures that any region predicted as fake by either model is retained. This strategy significantly boosts both the F1-score (to 0.7598) and IoU (to 0.6657), highlighting the complementary strengths of the two models—local detail sensitivity and global coverage.

LFDL + MITL + MDMF further applies post-processing to refine the fused mask using morphological operations (e.g., dilation, erosion). This enhances mask smoothness and removes small noisy regions, achieving the highest performance with a Final Score of 0.8150, F1-score of 0.7759, and IoU of 0.6902.

In summary, the LFDL module offers precise local forgery detection, while the MDMF module complements it by providing global contextual information. Their fusion—especially through naive logical integration and morphological refinement—yields substantial gains in both detection and localization, underscoring the value of combining local and global cues.

\subsection{Visualization}

\begin{figure}[!t]
\centering
\centerline{\includegraphics[width=0.8\columnwidth]{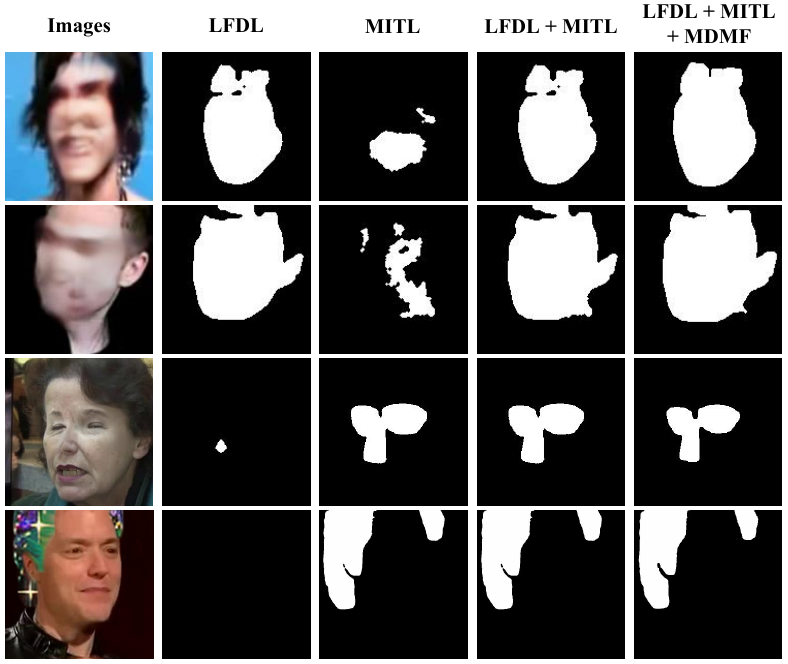}}
\caption{Visualization results of our methods. We apply dilation to LFDL masks and erosion to MITL masks, then combine them to achieve precise and coherent forgery localization.}
\label{fig_vis_result}
\end{figure}

As shown in Figure \ref{fig_vis_result}, our method achieves precise forgery localization by combining the strengths of LFDL and MITL. The LFDL module accurately detects facial manipulations but suffers from irregular edges and fragmented regions, while the MITL module effectively identifies peripheral artifacts (e.g., hair region anomalies) yet tends to over-expand boundaries. To address these issues, we innovatively adopt MDMF, which performs dilation to LFDL masks to smooth fragmented areas and erosion to MITL masks to suppress boundary over-extension, followed by an intersection operation for final localization. Experimental results demonstrate that this strategy significantly improves boundary precision and structural coherence of the detection masks.

\section{Conclusion}
In this work, we address the challenging task of deepfake localization by proposing a dual-branch framework that separately models local and global manipulation cues. 
While existing methods often overlook the importance of semantic-aware localization and suffer from ineffective fusion strategies, our approach explicitly leverages both fine-grained local details and global semantic context. 
By applying morphological fusion to the independently predicted masks, we suppress noisy activations and improve spatial coherence. 
Extensive experiments demonstrate the effectiveness of our method, showing that it achieves more accurate and robust localization results compared to existing approaches. 
This work highlights the importance of multi-perspective modeling and adaptive fusion in advancing the state-of-the-art in deepfake forensics.

\bibliographystyle{named}
\bibliography{main}

\end{document}